\documentclass[runningheads]{llncs}

\usepackage[dvipsnames]{xcolor}
\usepackage{latexsym}
\usepackage{amsmath}
\usepackage{svg}
\usepackage{booktabs}
\usepackage{multirow} 

\usepackage[T1]{fontenc}
\usepackage[utf8]{inputenc}
\usepackage{microtype}
\usepackage{inconsolata}

\usepackage{tikz}
\usetikzlibrary{positioning, shapes, calc, fit, decorations.pathmorphing, decorations.shapes}
\tikzset{>=latex}

\usepackage{pgf} 
\usepackage{colortbl}

\newcommand{\gradientcell}[7]{%
    \ifdimcomp{#1pt}{>}{#3 pt}{\cellcolor{#5!100.0!#4!#6}#7}{%
    \ifdimcomp{#1pt}{<}{#2 pt}{\cellcolor{#5!0.0!#4!#6}#7}{%
        \pgfmathparse{int(round(100*(#1/(#3-#2))-(#2 *(100/(#3-#2)))))}%
        \xdef\tempa{\pgfmathresult}%
        \cellcolor{#5!\tempa!#4!#6}#7%
}}}

\newcommand{\R}[2]{\gradientcell{#1}{51}{75}{Yellow}{Orange}{30}{#2}}

\DeclareMathOperator{\Bottle}{Bottleneck}
\DeclareMathOperator{\mse}{MSE}
\DeclareMathOperator{\mean}{mean}
\DeclareMathOperator{\softmax}{softmax}

\begin{document}
\title{Investigating the Effect of Parallel Data in the \\ Cross-Lingual Transfer for Vision-Language Encoders}
\titlerunning{Investigating Cross-Lingual Transfer for VL Encoders}


\author{Andrei-Alexandru Manea \and Jindřich Libovický}

\institute{Charles University, Faculty of Mathematics and Physics \\ Institute of Formal and Applied Linguistics \\  V Holešovičkách 2, 180 00 Prague, Czech Republic \\ \email{\texttt{\{manea, libovicky\}@ufal.mff.cuni.cz}}}

\authorrunning{A. Manea and J. Libovický}
%
\maketitle              

\begin{abstract}
Most pre-trained Vision-Language (VL) models and training data for the downstream tasks are only available in English.
Therefore, multilingual VL tasks are solved using cross-lingual transfer: fine-tune a multilingual pre-trained model or transfer the text encoder using parallel data. 
We study the alternative approach: transferring an already trained encoder using parallel data.
We investigate the effect of parallel data: domain and the number of languages, which were out of focus in previous work.
Our results show that even machine-translated task data are the best on average, caption-like authentic parallel data outperformed it in some languages.
Further, we show that most languages benefit from multilingual training.

\end{abstract}

\section{Introduction}

Most Vision-Language (VL) models suitable for multimodal classification consist of a language and a vision encoder \cite{vedaldi_uniter_2020,kwon_masked_2023,xu_bridgetower_2023}. They are mostly pre-trained in English, and most data for downstream task training only exists in English. Several multilingual models, usable for cross-lingual transfer, also exist \cite{ni_m3p_2021,zeng-etal-2023-cross}. 
A common approach is to pre-train a multilingual VL model with a multilingual text encoder, fine-tune it on English data, and test it on target languages (zero-shot cross-lingual transfer; \cite{pires-etal-2019-multilingual,conneau-etal-2020-unsupervised}). This carries the risk of forgetting other languages \cite{vu-etal-2022-overcoming}. An alternative is to fine-tune an English VL model for a downstream task, replace the English text encoder with a multilingual one, and fine-tune on parallel data \cite{karoui-etal-2023-stop}.

Although these methods perform well on standard benchmarks, previous work left many aspects unexplored. In this paper, we focus on the second approach using parallel data, and we address two questions: First, we examine what types of parallel data are most effective for cross-lingual transfer in VL models. Second, we analyze the impact on different languages, comparing independent encoders for individual languages, a single encoder for multiple languages, and zero-shot language transfer.
Building on CliCoTea: \cite{karoui-etal-2023-stop}, we propose a methodology using subword-aligned parallel data to fine-tune a multilingual text encoder, replacing the original text encoder in the BridgeTower \cite{xu_bridgetower_2023} VL encoder (\S~\ref{sec:method}).

We conduct experiments on natural language-vision reasoning (NLVR) and test using MARVL \cite{liu-etal-2021-visually} and M5-VGR \cite{schneider-sitaram-2024-m5}, datasets which contain images and sentences annotated by native speakers. Moreover, we repeated the setup with the XVNLI dataset \cite{bugliarello2022iglue}, considering English and non-English languages. More details are in \S~\ref{sec:datasets}.

First, we compare English-only, task MT, caption MT, generic, and caption-like in-domain parallel data (\S~\ref{subsec:cap_class}). Second, we evaluate the performance of bilingual and multilingual cross-lingual transfer, investigating the effects of incorporating additional languages into the parallel data (\S~\ref{subsec:more_languges}). 

Results (\S~\ref{sec:results}) show that caption-like parallel data surpass MT data only in a few cases. Moreover, using more languages for cross-lingual transfer, on average, leads to better results.

\begin{figure}[t]
\centering\scalebox{.75}{\begin{tikzpicture}[
        mpart/.style={draw, font=\footnotesize, align=center, rounded corners=1mm, inner sep=5pt},
        mbox/.style={draw, rounded corners=1mm, inner sep=5pt},
        databox/.style={draw, cylinder, shape aspect=.5, font=\footnotesize, shape border rotate=90, align=center, cylinder uses custom fill, cylinder body fill=Gray!20, cylinder end fill =Blue!20}]
    \begin{scope}[local bounding box=pretrained]
    \node[mpart, text width=7mm, fill=Red!20] (img1) {Img. enc.};
    \node[mpart,below=2pt of img1, text width=7mm, fill=Blue!20] (txt1) {Text enc.};
    \node[mpart,below right= -20pt and 5pt of img1, fill=Plum!30] (cm1) {\rotatebox{-90}{{X-modal}}};

    \draw[->] (img1) -- (cm1);
    \draw[->] (txt1) -- (cm1);

    \node[mbox, fit=(img1) (txt1) (cm1)] {};
    \end{scope}

    \node[above=1pt of pretrained] {Pre-trained model};

    \begin{scope}[
        local bounding box=taskData,
        xshift=35mm, yshift=-26mm]

        \node[databox, text width=7mm, text height=3mm] (txtdata) {Text data};

        \node[databox, text width=7mm, text height=3mm, left=5pt of txtdata, cylinder end fill =Red!20] (imgdata) {Img. data};
        
    \end{scope}

    \node[left=3pt of taskData, text width=30mm, align=right] {Multimodal task data};
    
    \begin{scope}[
        local bounding box=finetuned,
        xshift=50mm]
        
        \node[mpart, text width=7mm, fill=Red!20] (img2) {Img. enc.};
        \node[mpart,below=2pt of img2, text width=7mm, fill=Blue!20] (txt2) {Text enc.};
        \node[mpart,below right= -20pt and 5pt of img2, fill=Plum!30] (cm2) {\rotatebox{-90}{{X-modal}}};
    
        \draw[->] (img2) -- (cm2);
        \draw[->] (txt2) -- (cm2);

        \node[mbox, fit=(img2) (txt2) (cm2)] {};

    \end{scope}

    \node[above=1pt of finetuned] {Task-specific model};

    \draw[->] (pretrained) -- (finetuned) node[midway, draw, rounded corners=1mm, fill=Yellow!20] (ft) {\bf Finetuning};
    \draw[->] (imgdata) to[out=90, in=240] (ft);
    \draw[->] (txtdata) to[out=90, in=240] (ft);

    \begin{scope}[
        local bounding box=parallel,
        shift={($(taskData.west)+(82mm,-5mm)$)}]

        \node[databox, text width=12mm, text height=1.5mm, cylinder end fill = Blue!50] (bitext) {Bilingual text};
       
    \end{scope}

    \node[right=3pt of parallel, text width=20mm] {Caption-like parallel data};

    \begin{scope}[
        local bounding box=final,
        xshift=120mm]
        
        \node[mpart, text width=7mm, fill=Red!20] (img3) {Img. enc.};
        \node[mpart,below left=2pt and -30pt of img3, text width=11mm, fill=Blue!50] (txt3) {Muling. text enc.};
        \node[mpart,below right= -20pt and 5pt of img3, fill=Plum!30] (cm3) {\rotatebox{-90}{{X-modal}}};
    
        \draw[->] (img3) -- (cm3);
        \draw[->] (txt3) -- (cm3);

        \node[mbox, fit=(img3) (txt3) (cm3)] {};

    \end{scope}

    \node[above=1pt of final] {Final model};

    \draw[->] (finetuned) -- (final);
    
    \begin{scope}[
        local bounding box=transfer,
        xshift=82mm,
        yshift=0mm]
        \node[mpart,text width=11mm, fill=Blue!50] (txtMulti) {Muling. text enc.};
        \node[mpart,below=10pt of txtMulti, text width=7mm, fill=Blue!30] (txtMono) {Text enc.};

        \node[mbox, fit=(txtMono) (txtMulti), line width=2pt, fill=Yellow!20, , inner sep=8pt] {};

        \node[mpart,text width=11mm, fill=Blue!50] (txtMulti) {Muling. text enc.};
        \node[mpart,below=10pt of txtMulti, text width=7mm, fill=Blue!20] (txtMono) {Text enc.};
        \draw[thick, double] (txtMulti) -- (txtMono);
        
    \end{scope}
    \node[above=1pt of transfer] {\bf Encoder Transfer};

    \draw[->] (txtdata) to[out=0,in=240] (transfer);
    \draw[->] (bitext) to[out=180,in=270] (transfer);

    \draw[->,line width=2pt, dotted, Gray] (txt2) to[out=300, in=180] (txtMono);
    \draw[->,line width=2pt, dotted, Gray] (txtMulti) to[out=320, in=180] (txt3);

\end{tikzpicture}}
\caption{Overall scheme of the proposed approach. The VL model is depicted as a box containing both the image and the text encoder, combined with the violet cross-modal encoder. The VL model is modified in two phases: (1)~\textbf{Fine-tune} the entire model using task-specific data and (2)~\textbf{Encoder Transfer}, which transfers the English text encoder capabilities into the multilingual one, using English task-specific data and bilingual caption-like samples; the resulting text encoder is placed in the final state.}
\label{fig:OverallScheme}
\end{figure}
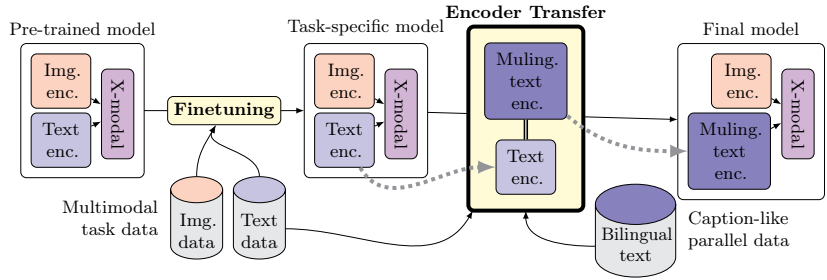

\section{Related Work}

Classification tasks combining language and vision are often approached by fine-tuning pre-trained VL encoders. These models typically include a pre-trained vision encoder, a pre-trained text encoder, and modules combining modalities, which are usually trained from scratch.
Most VL models, such as LXMert \cite{tan-bansal-2019-lxmert}, MaskVLM \cite{kwon_masked_2023}, ALBEF \cite{li_align_2021}, BridgeTower \cite{xu_bridgetower_2023}, and X2-VLM \cite{zeng_x2-vlm_2023}, are trained on English data. Task-specific fine-tuning data is also mostly in English. Extending VL models to other languages, therefore, relies on cross-lingual transfer.

The two main approaches to cross-lingual transfer with VL models are: \emph{pre-train a multilingual encoder} for fine-tuning with English or multilingual data, and \emph{transfer a task-specific encoder} using parallel data. 

\paragraph{Multilingual Pre-training.} 

VL models are usually pre-trained using variants of Masked Language Modeling (MLM; \cite{devlin-etal-2019-bert}) extended to the vision modality as Masked Region Modeling (MRM). For instance, MaskVLM \cite{kwon_masked_2023} randomly masks both modalities, whereas ALBEF \cite{li_align_2021} only masks one modality, which is reconstructed using the other. Most VL encoders fine-tune existing text and vision encoders and only train the cross-modal parts of the model.

This pre-training can be extended for multiple languages using a pre-trained multilingual text encoder, such as XLM-R \cite{conneau-etal-2020-unsupervised}. In this case, MLM can be applied on monolingual text \cite{zhou_uc2_2021,li_align_2021,zeng_x2-vlm_2023}, parallel data \cite{liu2021visually}, or code-switched texts \cite{ni_m3p_2021}. 

\paragraph{Transfer with parallel data.} 

Parallel data are often used for cross-lingual transfer in text-only encoders at the word \cite{cao2020multilingual,wu-dredze-2020-explicit} and sentence levels \cite{reimers-gurevych-2020-making,feng-etal-2022-language}, but they are less common with VL models.

mCLIP \cite{carlsson-etal-2022-cross} extends CLIP \cite{radford2021learning} to multiple languages by replacing the CLIP text encoder with mBERT \cite{devlin-etal-2019-bert}, fine-tuned to produce similar embeddings to CLIP using Mean Squared Error ($\mse$). This approach works only for single-vector sentence embeddings and relies on machine-translated image captions.

Methods aligning hidden states on the subword level \cite{cao2020multilingual} often depend on external unsupervised word aligners. In VL, the only work we found is CliCoTea \cite{karoui-etal-2023-stop}, which, like mCLIP, uses data from Google Translate. It uses Awesome Align \cite{dou2021word} for word alignment and fine-tunes XLM-R to mimic the English text encoder's states. However, because Awesome Align aligns words, not subwords, many tokens remain unaligned and missing from the loss function. CliCoTea is theoretically multilingual but reports results only for bilingual transfer.

It is unclear how sensitive VL models are to parallel data quality, whether authentic data could replace machine-translated, or to what extent these methods are affected by the curse of multilinguality \cite{pfeiffer-etal-2022-lifting}. In this paper, we fill this gap.
 
\section{Methodology}
\label{sec:method}

Similarly to CliCoTea \cite{karoui-etal-2023-stop},
we transfer a model fine-tuned for a downstream task to different languages by replacing the original monolingual English encoder with a multilingual encoder trained to mimic the monolingual encoder.
Figure~\ref{fig:OverallScheme} describes the overview of the method.

We use textual data, both monolingual and parallel, to do the encoder transfer and use subword alignment. 
For two subword sequences of lengths $T_1$ and $T_2$, we define the alignment $A \in \{0, \ldots, T_1\} \times \{0, \ldots, T_2\}$, as a set of pairs of indices of aligned subwords, and the alignment loss
\begin{align}
    L^{(k)}_{align} = \frac{1}{|A|} \sum_{(i,j) \in A} (h^{(i)}_k - g^{(j)}_k )^2
\end{align}
where $h^{(i)}_k$ is the hidden state of the $i$-th token in the $k$-layer of the monolingual encoder and $g^{(j)}_k$ is the hidden state of the $j$-th token in the $k$-layer of the multilingual encoder.

To not exclude hidden states of subwords not covered by the alignment, we further want the mean-pooled hidden states on each layer to match.
Similarly to mCLIP \cite{carlsson-etal-2022-cross}, we minimize the distance between the pooled states:
\begin{align}
    L^{(k)}_{mean} = \mse(\mean_j(h^{(j)}_k), \mean_j(g^{(j)}_k))
\end{align}

To aggregate this across all targeted layers, our final loss becomes:
\begin{align}
    L = \mean_k \left( L^{(k)}_{align} + L^{(k)}_{mean} \right)
\end{align}

To allow for more flexibility, we do not work directly with the hidden state of the multilingual encoder but with a non-linear feed-forward layer with GELU non-linearity over a linear combination of the encoder's hidden states:
    $g^*_i = \Bottle\left(\sum^{K-1}_{k=0} a_k g_k\right)$
where $a = \softmax(p)$, and $p$ is a learnable vector. To validate this method, we perform a few ablation experiments (\S~\ref{sec:app:architecture}).

\section{Experiments}

Our experiments use two data selection methods: \S~\ref{subsec:cap_class} and \S~\ref{subsec:more_languges}. The following section describes the setup that is common for both experiments.

\subsection{Encoder Transfer Setup}

\paragraph{VL encoder.} We work with BridgeTower \cite{xu_bridgetower_2023}, a VL model that uses CLIP-ViT visual encoder \cite{radford_learning_2021} for image encoding and RoBERTa Large \cite{liu_roberta_2019} encoding.
We aim to replace RoBERTa with the multilingual encoder XLM-R Large \cite{conneau-etal-2020-unsupervised}.
BridgeTower uses the last six layers for cross-modal encoding.
Therefore, we fine-tune the last six layers of XLM-R to output the same hidden states as English RoBERTa.

\paragraph{Downstream task data.}
We conduct our study on natural language-vision reasoning on NLVR2 \cite{suhr2019corpus}, MARVL \cite{liu2021visually}, M5-VGR \cite{schneider-sitaram-2024-m5} and XVNLI \cite{bugliarello2022iglue} with its English subset, which we call VNLI. 


%

\paragraph{Text data.}
In our transfer experiments, we use 2 datasets: English and multilingual.
For the English one, we uniformly sampled 200k sentences from training sets of tasks in the IGLUE benchmark \cite{bugliarello2022iglue}. The multilingual parallel data differ for different experiments and are described in more detail in the following section.

\paragraph{Text preprocessing.}
Our method requires subword-level alignment of the texts fed into the text encoders.
We use Eflomal \cite{ostling2016efficient} with the grow-diagonal heuristic to obtain the alignment.
RoBERTa and XLM-R use different subword tokenizers; therefore, we need to get the subword alignment for both English and bilingual texts. Since the parallel data is extracted from a bigger pool, as mentioned in \S~\ref{subsec:cap_class}, we first computed the alignments inside the pool. We used them as priors for the alignments of the sentences, which we considered for the transfer.

\subsection{Experiment 1: Parallel Data Type}
\label{subsec:cap_class}

In this experiment, we compare four data strategies for the transfer: (1) English only, using only the English dataset, Machine Translated samples from NLVR2 (2) or MSCOCO (3), obtained with Google Translate\footnote{https://pypi.org/project/googletrans/}, (4) generic parallel data, and (5) caption-like parallel data. 

We use OPUS-100 \cite{zhang-etal-2020-improving,tiedemann-2012-parallel} as our main source of authentic parallel data. For Swahili, which is missing in OPUS-100, we used parallel data from the MultiHPLT 1.1 dataset\footnote{https://hplt-project.org/datasets/v1.1}. For Berber and Filipino, which are also missing, we do not collect any additional data; hence, the results are zero-shot. From this corpus, we filtered 5k in-domain sentence pairs containing each language from the multilingual set and English. We trained a classifier to find English sentences resembling captions from the VL datasets by using sentences from the MSCOCO as positive examples and OPUS-100 as negative examples. More details are in the Appendix \ref{sec:cap_class}.

\begin{table}[t]
\caption{Results of the cross-lingual transfer with different types of parallel data for MARVL, M5VGR, and XVNLI. 
 M5B-VGR results per language are Table~\ref{tab_ablat_1_mgb_vgr} in the appendix.
}\label{tab_ablat_1}%
\centering\scalebox{0.85}{
\setlength{\tabcolsep}{1.5pt}
\begin{tabular}{ll c@{\hskip 8pt}ccccc@{\hskip 8pt}c@{\hskip 8pt}c@{\hskip 8pt}c@{\hskip 8pt}cccc@{\hskip 8pt}c}
\toprule

\multicolumn{2}{l}{Data} & \textls[-120]{NLVR2} & \multicolumn{6}{c}{MARVL} & \textls[-140]{M5-VGR} & \textls[-100]{VNLI} & \multicolumn{5}{c}{XVNLI} \\ \cmidrule(l{-1pt}r{7pt}){3-3} \cmidrule(l{-1pt}r{7pt}){4-9} \cmidrule(l{-1pt}r{7pt}){10-10} \cmidrule(l{-2pt}r){11-11} \cmidrule(l{3pt}r{3pt}){12-16}
 & & en  & tr & sw & zh & id & ta & avg & avg & en & ar & es & fr & ru & avg \\
\midrule
\multicolumn{2}{l}{Majority Class} & \R{49.0}{49.0} & \R{50.2}{50.2} & \R{50.6}{50.6} & \R{50.2}{50.2} & \R{50.0}{50.0} & \R{50.4}{50.4} & \R{50.3}{50.3} & \R{58.5}{58.5} & \R{33.9}{33.9} & \R{33.9}{33.9} & \R{33.9}{33.9} & \R{33.9}{33.9} & \R{33.9}{33.9} & \R{33.9}{33.9} \\
\multicolumn{2}{l}{English only} & \R{81.8}{81.8} &\R{51.1}{51.1} &\R{50.5}{50.5} &\R{59.1}{59.1} &\R{60.3}{60.3} &\R{57.6}{57.6} &\R{55.7}{55.7} & \R{46.8}{46.8} & \R{85.2}{\textbf{85.2}} & \R{47.9}{47.9} & \R{47.6}{47.6} & \R{54.6}{54.6} & \R{61.7}{61.7} & \R{52.9}{52.9} \\ \midrule
\multirow{3}{*}{\rotatebox[]{90}{Parallel}}
& Task MT & \R{81.8}{81.8} & \R{71.9}{\textbf{71.9}} & \R{68.1}{\textbf{68.1}} & \R{73.7}{\textbf{73.7}} & \R{68.6}{68.6} & \R{65.9}{\textbf{65.9}} & \R{69.5}{\textbf{69.5}} & \R{54.1}{\textbf{54.1}} & \R{85.1}{85.1} & \R{76.5}{\textbf{76.5}} & \R{77.1}{\textbf{77.1}} & \R{78.3}{\textbf{78.3}} & \R{77.8}{\textbf{77.8}} & \R{77.4}{\textbf{77.4}} \\
& Caption MT & \R{81.8}{81.8} & \R{65.5}{65.5} & \R{65.4}{65.4} & \R{60.8}{60.8} & \R{61.9}{61.9} & \R{62.6}{62.6} & \R{63.3}{63.3} & \R{54.1}{\textbf{54.1}} & \R{85.1}{85.1} & \R{71.2}{71.2} & \R{75.9}{75.9} & \R{76.9}{76.9} & \R{75.6}{75.6} & \R{74.9}{74.9} \\
& Generic & \R{81.8}{81.8} &\R{59.2}{59.2} &\R{59.3}{59.3} &\R{63.0}{63.0} &\R{66.2}{66.2} &\R{59.5}{59.5} &\R{61.4}{61.4} & \R{50.9}{50.9} & \R{84.8}{84.8} & \R{61.8}{61.8} & \R{69.5}{69.5} & \R{70.0}{70.0} & \R{66.6}{66.6} & \R{66.9}{66.9} \\
& Caption-like & \R{81.8}{81.8} &\R{67.9}{67.9} & \R{65.9}{65.9} & \R{68.5}{68.5} & \R{69.1}{\textbf{69.1}} & \R{60.9}{60.9} & \R{66.3}{66.3} &  \R{53.2}{53.2} & \R{84.8}{84.8} & \R{70.9}{70.9} & \R{72.3}{72.3} & \R{75.7}{75.7} & \R{71.3}{71.3} & \R{72.6}{72.6} \\
\bottomrule
\end{tabular}
}

\end{table}

\subsection{Experiment 2: Number of languages}
\label{subsec:more_languges}

In this experiment, we investigate the effect of adding more languages in the cross-signal transfer. We start with bilingual encoders similar to CliCoTea: \cite{karoui-etal-2023-stop} and prepare five encoders for each MARVL language paired with English.
We use 5k caption-like sentence pairs for each language and train for 10 epochs. We compare this with the encoder trained in all languages in 50 epochs.

In the next step, we add more languages to the training and use 5k parallel sentence pairs for each bilingual pair of languages. We run experiments with 10, 20, 30, and 40 languages. We draw the additional languages from the OPUS-100 dataset. The list of languages is in Appendix~\ref{sec:appendix}.

\begin{table}
\caption{Results of the cross-lingual transfer with bilingual data. The \textit{Multiling.} line represents \textit{Caption-like par.} results from Table~\ref{tab_ablat_1}. The \textit{Bilingual} line shows the results of bilingual transfer. Zero-shot results of bilingual transfer into other languages are in Table~\ref{tab_ablat_2_detailed}.}\label{tab_ablat_2}

\centering

\scalebox{0.8}{
\setlength{\tabcolsep}{3.3pt}
\begin{tabular}{l c@{\hskip 9pt}ccccc@{\hskip 9pt}c}
\toprule

Data & \textls[-120]{NLVR2} & \multicolumn{6}{c}{MARVL} \\ \cmidrule(lr{9pt}){2-2}  \cmidrule(lr){3-8}
 & en  & tr & sw & zh & id & ta & avg \\
\midrule
Biling. 5k & --- &\R{69.6}{69.6} &\R{64.2}{64.2} &\R{68.0}{68.0} &\R{67.1}{67.1} &\R{60.5}{60.5} &\R{65.9}{65.9} \\
Multi. 25k &\R{81.8}{81.8} &\R{67.9}{67.9} &\R{65.9}{65.9} &\R{68.5}{68.5} & \R{69.1}{\textbf{69.1}} &\R{60.9}{60.9} &\R{66.3}{66.3} \\
Biling. 25k & --- & \R{73.2}{\textbf{73.2}} & \R{66.9}{\textbf{66.9}} & \R{71.2}{\textbf{71.2}} &\R{67.9}{67.9} & \R{61.2}{\textbf{61.2}} & \R{68.1}{\textbf{68.1}} \\
\bottomrule
\end{tabular}
}

\end{table}


\begin{table}
\caption{Results of the cross-lingual transfer by including more languages. The \textit{5} language line represents \textit{Caption-like par.} results from Table~\ref{tab_ablat_1}. More details about the language selected for each experiment are in the Appendix~\ref{sec:appendix}.}\label{tab_ablat_3}

\centering
\scalebox{0.8}{
\setlength{\tabcolsep}{3.3pt}
\begin{tabular}{l c@{\hskip 10pt}ccccc@{\hskip 10pt}c}
\toprule

\# lang & \textls[-120]{NLVR2} & \multicolumn{6}{c}{MARVL} \\ \cmidrule(lr{9pt}){2-2}  \cmidrule(lr){3-8}
 & en  & tr & sw & zh & id & ta & avg \\
\midrule
\ 5 & \R{81.8}{\textbf{81.8}} &\R{67.9}{67.9} &\R{65.9}{65.9} &\R{68.5}{68.5} & \R{69.1}{\textbf{69.1}} &\R{60.9}{60.9} &\R{66.3}{66.3} \\
10 & \R{81.8}{\textbf{81.8}} &\R{68.2}{68.2} &\R{66.2}{66.2} & \R{69.6}{\textbf{69.6}} &\R{68.4}{68.4} &\R{60.7}{60.7} &\R{66.5}{66.5} \\
20 &\R{81.7}{81.7} & \R{70.4}{\textbf{70.4}} & \R{66.9}{\textbf{66.9}} &\R{68.2}{68.2} &\R{68.4}{68.4} &\R{60.4}{60.4} & \R{66.7}{\textbf{66.7}} \\
30 &\R{81.7}{81.7} &\R{69.7}{69.7} &\R{64.9}{64.9} &\R{68.5}{68.5} &\R{69.0}{69.0} &\R{61.7}{61.7} & \R{66.7}{\textbf{66.7}} \\
40 &\R{81.7}{81.7} &\R{69.9}{69.9} &\R{64.8}{64.8} &\R{67.4}{67.4} &\R{67.6}{67.6} & \R{61.9}{\textbf{61.9}} &\R{66.3}{66.3} \\
\bottomrule
\end{tabular}
}
\end{table}

\section{Results}
\label{sec:results}

Table~\ref{tab_ablat_1} and Table~\ref{tab_ablat_1_mgb_vgr} present the accuracy for each dataset.
We include English accuracy to track if our models suffer from catastrophic forgetting.
Despite using a multilingual text encoder, the English-only fine-tuning leads to results that are not far from the random baseline.
On average, we reached our best results with Task MT data, except for Indonesian (MARVL) and German, Filipino, Hindi, Swahili, and Thai (M5-VGR). 
%

Table~\ref{tab_ablat_2} compares the performance of bilingual models on MARVL, trained on 5k or 25k caption-like samples, to that of the multilingual model (5k samples per language).
%
%
The average results from Table~\ref{tab_ablat_2} show that multilingual data improves performance in the very low-resource scenario. However, except for Indonesian, using the same amount of bilingual data outperforms the multilingual mix. In this way, we match the results of CliCoTea: \cite{karoui-etal-2023-stop} with only a fraction of caption-like parallel data. For more comparison with related work, see Appendix~\ref{sec:app:related}.

Table~\ref{tab_ablat_3} shows the results of experiments with adding more languages into the training, preserving 5k caption-like samples for each language. We observe a synergy effect of having more languages, up to 15 additional languages. As the number increases further, the accuracy slightly diminishes.

\section{Conclusions}

We conducted two experiments to better understand the role of parallel data in the cross-lingual transfer of VL models.
%
In our first experiment, we show that the common strategy (mCLIP \cite{carlsson-etal-2022-cross}; CliCoTea \cite{karoui-etal-2023-stop}) is not the only solution, and caption-like parallel data can lead to better results. 
Our second experiment shows that, on average, multilingual training outperforms bilingual transfer in a limited scenario, but not as much as using more bilingual samples.
The language synergy in the multilingual setup goes beyond five test languages, reaching the best average results with 20 languages, but diminishes with more languages.





\section{Acknowledgment}

We thank Jindřich Helcl, Simone Balloccu, and Gianluca Vico for comments on the draft of the paper.
This research was supported by the Charles University project PRIMUS/23/SCI/023 and SVV project number 260 821.

%
%
%
\bibliographystyle{splncs04}
\bibliography{biblio}

\appendix

\section{Caption Classifier}
\label{sec:cap_class}
To train the classifier, we create an English dataset with 500k sentences: negative examples from OPUS-100 and positive examples from COCO captions. We fine-tuned RoBERTa  \cite{liu_roberta_2019} with a batch size of 128, learning rate of $2 \cdot 10^{-5}$ , and linear weight decay of $10^{-2}$ for 16k steps, reaching validation F$_1$ score $99.9\%$ on a held-out set. 
We use the trained classifier to score English sentences from OPUS-100 and select the top 5k sentences for each language in MARVL. Together with their corresponding bilingual pairs (up to 450 tokens) using BERTScore \cite{zhang2020bertscore}, we select the top 5k pairs with the most similar meaning.

We train the multilingual model on the obtained parallel sentences, with the batch size $128$, learning rate $10^{-4}$, linear weight decay of $10^{-2}$, warmup ratio of $10\%$ for $50$ epochs.

\section{Dataset Description}
\label{sec:datasets}

We evaluate our model on three multilingual visual-language reasoning datasets:

\paragraph{MARVL.} A benchmark \cite{liu-etal-2021-visually} with image-text pairs, where models determine whether a textual hypothesis is true or false. Unlike other datasets that extend from an English-based foundation, MARVL was independently developed for five languages, focusing on cultural aspects: Turkish, Swahili, Chinese, Indonesian, and Tamil.

\paragraph{M5-VGR.} Following the same structure as MARVL, but covering 12 more diverse languages: Amharic, Berber, Bengali, German, Filipino, Hausa, Hindi, Russian, Swahili, Thai, and Zulu. 

\paragraph{XVNLI.}: A dataset that extends natural language inference into the multimodal and multilingual world, covering five languages (English, Arabic, French, Russian, and Spanish). The task involves classifying relationships between an image and two textual hypotheses (entailment, neutral, or contradiction).

In our experiments, we exclude the English subset from M5-VGR.
\section{Detailed Results}
\label{sec:appendix}


The sets from Table~\ref{tab_ablat_3} contain, in cascade, the following languages:
\begin{itemize}
    \item 10 languages -- additionally, Arabic, Bengali, Bulgarian, Danish, Estonian
    \item 20 languages -- all the languages from IGLUE; additionally English (identical pairs), German, Greek, French, Japanese, Korean, Portuguese, Russian, Spanish, Vietnamese 
    \item 30 languages -- additionally, Czech, Romanian, Welsh, Amharic, Igbo, Yoruba, Malagasy, Punjabi, Pashto, Thai
    \item 40 languages -- additionally, Assamese, Gujarati, Hindi, Kannada, Malayalam, Marathi, Oriya, Sinhalese, Persian, Hebrew
\end{itemize}





\subsection{Architecture Ablation Study}
\label{sec:app:architecture}

We conducted 4 additional experiments on the task MT parallel data to assess the effect of differences of our model compared to CliCoTea: \cite{karoui-etal-2023-stop}. We evaluate the performance only on the MARVL dataset, with results shown in Table~\ref{tab_ablat_architecture}. On average, our method, \textit{Exp. 0} performs the best.

The weighted layers represent the mentioned linear combination $\sum^{K-1}_{k=0} a_k g_k$, while the case of using the last layer only is the raw value $g_{K-1}$. The rule applies to all the layers that are inputted to the cross-modal encoder.

The $\Bottle$ is a simple small sequence of (a) linear downsampling to 4 times less the hidden size, followed by GeLU activation and Layer Normalization, and (b) upsampling back to the initial size. In these ablations, we replace this module with a simple linear projection (FFN) or remove it (identity projection).

\subsection{Comparison with Related Work}
\label{sec:app:related}

Table~\ref{tab_macro_compared} compares our model performance and the number of parameters to existing models. Although results are not directly comparable, we match CliCoTea's \cite{karoui-etal-2023-stop} performance with only 25k examples, a fraction of their set. Notably, our best model and CliCoTea achieve strong performance with under 1B parameters, whereas mBLIP \cite{geigle-etal-2024-mblip}, with the highest performance, consists of a few billion parameters.

\begin{table}
\centering
\caption{Results of the cross-lingual transfer with 5k (left side) and 25K (right side) bilingual data. The \textit{Multiling.} line represents \textit{Captions-like.} results from Table~\ref{tab_ablat_1}, and the native is the diagonal of the table, i.e. the results of each bilingual encoder on the selected language from MARVL.}

\scalebox{0.8}{
\setlength{\tabcolsep}{3.3pt}
\begin{tabular}{l c@{\hskip 9pt}ccccc@{\hskip 9pt}c@{\hskip 10pt} c@{\hskip 10pt}ccccc@{\hskip 10pt}c}
\toprule
Data & \textls[-120]{NLVR2} & \multicolumn{6}{c}{MARVL} & \textls[-120]{NLVR2} & \multicolumn{6}{c}{MARVL} \\ \cmidrule(l{-1pt}r{7pt}){2-2}  \cmidrule(l{-1pt}r{7pt}){3-8} \cmidrule(l{-1pt}r{7pt}){9-9} \cmidrule(l{-1pt}r{7pt}){10-15}
 & en  & tr & sw & zh & id & ta & avg & en  & tr & sw & zh & id & ta & avg \\
\midrule
Multiling. &\R{81.8}{81.8} &\R{67.9}{67.9} &\R{65.9}{\textbf{65.9}} &\R{68.5}{\textbf{68.5}} & \R{69.1}{\textbf{69.1}} &\R{60.9}{\textbf{60.9}} &\R{66.3}{\textbf{66.3}} & 
\R{81.8}{81.8} &\R{67.9}{67.9} &\R{65.9}{65.9} &\R{68.5}{68.5} & \R{69.1}{\textbf{69.1}} &\R{60.9}{60.9} &\R{66.3}{66.3} \\
Biling. & --- &\R{69.6}{\textbf{69.6}} &\R{64.2}{64.2} &\R{68.0}{68.0} &\R{67.1}{67.1} &\R{60.5}{60.5} &\R{65.9}{65.9} & 
--- & \R{73.2}{\textbf{73.2}} & \R{66.9}{\textbf{66.9}} & \R{71.2}{\textbf{71.2}} &\R{67.9}{67.9} & \R{61.2}{\textbf{61.2}} & \R{68.1}{\textbf{68.1}} \\
\midrule
tr\_en &\R{81.5}{81.5} & \R{69.6}{\textbf{69.6}} &\R{56.8}{56.8} &\R{59.1}{59.1} &\R{63.1}{63.1} &\R{56.0}{56.0} &\R{60.9}{60.9} & 
\R{81.8}{\textbf{81.8}} & \R{73.2}{\textbf{73.2}} &\R{54.8}{54.8} &\R{57.1}{57.1} &\R{62.7}{62.7} &\R{58.1}{58.1} & \R{61.3}{\textbf{61.3}} \\
sw\_en &\R{81.5}{81.5} &\R{57.1}{57.1} & \R{64.2}{\textbf{64.2}} &\R{61.1}{61.1} &\R{64.5}{64.5} &\R{59.7}{59.7} & \R{61.2}{\textbf{61.2}} & 
\R{81.8}{\textbf{81.8}} &\R{57.2}{57.2} & \R{66.9}{\textbf{66.9}} &\R{59.8}{59.8} &\R{62.6}{62.6} &\R{58.4}{58.4} &\R{60.9}{60.9} \\
zh\_en &\R{81.5}{81.5} &\R{59.7}{59.7} &\R{54.1}{54.1} & \R{68.0}{\textbf{68.0}} &\R{60.9}{60.9} & \R{60.9}{\textbf{60.9}} &\R{60.6}{60.6} &
\R{81.7}{81.7} &\R{56.4}{56.4} &\R{51.5}{51.5} & \R{71.2}{\textbf{71.2}} &\R{61.1}{61.1} &\R{58.9}{58.9} &\R{59.6}{59.6} \\
id\_en &\R{81.5}{81.5} &\R{57.6}{57.6} &\R{57.1}{57.1} &\R{63.1}{63.1} & \R{67.1}{\textbf{67.1}} &\R{59.4}{59.4} &\R{60.8}{60.8} & 
\R{81.6}{81.6} &\R{55.9}{55.9} &\R{52.7}{52.7} &\R{60.7}{60.7} & \R{67.9}{\textbf{67.9}} &\R{56.8}{56.8} &\R{58.7}{58.7} \\
ta\_en &\R{81.5}{81.5} &\R{61.3}{61.3} &\R{55.3}{55.3} &\R{56.9}{56.9} &\R{62.3}{62.3} &\R{60.5}{60.5} &\R{59.4}{59.4} &
\R{81.7}{81.7} &\R{55.6}{55.6} &\R{52.2}{52.2} &\R{54.5}{54.5} &\R{62.1}{62.1} & \R{61.2}{\textbf{61.2}} &\R{57.3}{57.3} \\
\bottomrule
\end{tabular}
}
\label{tab_ablat_2_detailed} 
\end{table}

\begin{table}
\centering
\caption{Ablation study of the proposed projection architecture. Experiment 0 is \textit{Task MT} from Table~\ref{tab_ablat_1}.}

\scalebox{0.8}{
\setlength{\tabcolsep}{3.3pt}
\begin{tabular}{l ccccccccc}

\toprule
& \multicolumn{3}{c}{Projection} & \multicolumn{5}{c}{MARVL} \\
\cmidrule(l{1pt}r{9pt}){2-4} \cmidrule(l{1pt}r{9pt}){5-10} 
Exp. & Loss & Input & Layer & tr & sw & zh & id & ta & avg \\
\toprule

0 & $L_{align} + L_{mean}$ & weighted layers & $\Bottle$ & \R{71.9}{71.9} & \R{68.1}{\textbf{68.1}} & \R{73.7}{\textbf{73.7}} & \R{68.6}{68.6} & \R{65.9}{65.9} & \R{69.5}{\textbf{69.5}} \\
1 & $L_{align}$ & weighted layers & $\Bottle$ & \R{72.3}{72.3} & \R{65.5}{65.5} & \R{71.3}{71.3} & \R{68.9}{68.9} & \R{66.3}{66.3} & \R{68.8}{68.8} \\
2 & $L_{align} + L_{mean}$ & weighted layers & FFN & \R{72.3}{72.3} & \R{66.5}{66.5} & \R{71.7}{71.7} & \R{68.3}{68.3} & \R{67.1}{\textbf{67.1}} & \R{69.1}{69.1} \\
3 & $L_{align} + L_{mean}$ & last layer only & $\Bottle$ & \R{72.6}{\textbf{72.6}} & \R{67.1}{67.1} & \R{72.2}{72.2} & \R{68.7}{68.7} & \R{64.0}{64.0} & \R{68.8}{68.8} \\
4 & $L_{align} + L_{mean}$ & last layer only & identity & \R{72.6}{\textbf{72.6}} & \R{67.1}{67.1} & \R{72.3}{72.3} & \R{69.5}{\textbf{69.5}} & \R{61.9}{61.9} & \R{68.6}{68.6} \\

\bottomrule
\end{tabular}
}
\label{tab_ablat_architecture}
\end{table}

\begin{table}
\caption{Results of the cross-lingual transfer with different types of parallel data for M5B-VGR.}
\label{tab_ablat_1_mgb_vgr}

\centering

\scalebox{0.8}{
\setlength{\tabcolsep}{3.25pt}
\begin{tabular}{ll c@{\hskip 9pt}ccccccccccc@{\hskip 9pt}c}
\toprule

\multicolumn{2}{l}{Data} & \textls[-120]{NLVR2} & \multicolumn{11}{c}{M5-VGR} \\ \cmidrule(lr{9pt}){3-3} \cmidrule(l{1pt}r{9pt}){4-15} 

 & & en & am & ber & bn & de & fil & ha & hi & ru & sw & th & zu & avg \\
\midrule
\multicolumn{2}{l}{Majority Class} & \R{49.0}{49.0} & \R{56.7}{56.7} & \R{50.0}{50.0} & \R{63.6}{63.6} & \R{54.2}{54.2} & \R{52.5}{52.5} & \R{41.7}{41.7} & \R{61.9}{61.9} & \R{59.2}{59.2} & \R{64.2}{64.2} & \R{64.2}{64.2} & \R{63.8}{63.8} & \R{58.5}{58.5} \\

\multicolumn{2}{l}{English only} & \R{81.8}{\textbf{81.8}} & \R{44.2}{44.2} & \R{50.0}{50.0} & \R{36.4}{36.4} & \R{44.2}{44.2} & \R{52.5}{52.5} & \R{42.5}{42.5} & \R{46.6}{46.6} & \R{42.5}{42.5} & \R{38.3}{38.3} & \R{54.2}{54.2} & \R{36.2}{36.2} & \R{46.8}{46.8} \\ \midrule
\multirow{3}{*}{\rotatebox[]{90}{Parallel}}
& Task MT & \R{81.7}{81.7} & \R{65.8}{\textbf{65.8}} & \R{50.8}{\textbf{50.8}} & \R{48.3}{\textbf{48.3}} & \R{42.5}{42.5} & \R{52.5}{52.5} & \R{56.7}{\textbf{56.7}} & \R{44.1}{44.1} & \R{55.8}{\textbf{55.8}} & \R{50.0}{50.0} & \R{57.5}{57.5} & \R{47.4}{47.4} & \R{54.1}{\textbf{54.1}} \\
& Caption MT & \R{81.7}{81.7} & \R{63.3}{63.3} & \R{48.3}{48.3} & \R{46.6}{46.6} & \R{48.3}{48.3} & \R{52.5}{52.5} & \R{55.8}{55.8} & \R{47.5}{47.5} & \R{52.5}{52.5} & \R{46.7}{46.7} & \R{61.7}{61.7} & \R{52.6}{\textbf{52.6}} & \R{54.1}{\textbf{54.1}} \\
& Generic & \R{81.5}{81.5} & \R{52.5}{52.5} & \R{50.0}{50.0} & \R{38.9}{38.9} & \R{49.2}{\textbf{49.2}} & \R{55.0}{55.0} & \R{53.4}{53.4} & \R{43.2}{43.2} & \R{50.8}{50.8} & \R{43.3}{43.3} & \R{52.5}{52.5} & \R{43.9}{43.9} & \R{50.9}{50.9} \\
& Caption-like & \R{81.6}{81.6} & \R{55.8}{55.8} & \R{50.0}{50.0} & \R{46.6}{46.6} & \R{49.2}{\textbf{49.2}} & \R{57.5}{\textbf{57.5}} & \R{44.2}{44.2} & \R{50.8}{\textbf{50.8}} & \R{46.7}{46.7} & \R{55.0}{\textbf{55.0}} & \R{64.2}{\textbf{64.2}} & \R{41.4}{41.4} & \R{53.2}{53.2} \\
\bottomrule
\end{tabular}
}
\end{table}

\begin{table}
\caption{Results of our cross-lingual transfer compared to other existing methods. Our results are a copy of the Table~\ref{tab_ablat_2}. }
\label{tab_macro_compared}

\centering
\scalebox{0.8}{
\setlength{\tabcolsep}{3.3pt}
\begin{tabular}{l cccccccc}
\toprule
\multicolumn{2}{c}{Model} & \# parameters & tr & sw & zh & id & ta & avg \\
\toprule
\textit{Model based on multilingual pre-training} \\
mUNITER  			& \multirow{2}{*}{\cite{liu-etal-2021-visually}} & 137M & 54.7 & 51.2 & 55.3 & 54.8 & 52.7 & 53.7 \\
xUNITER  			&  & 152M & 56.2 & 55.5 & 53.1 & 55.1 & 53.1 & 54.6 \\
UC2      			& \cite{zhou_uc2_2021} & 270M & 56.7 & 52.6 & 59.9 & 56.7 & 60.5 & 57.3 \\
M3P      			& \cite{ni_m3p_2021} & 377M & 56.8 & 55.7 & 55.0 & 56.5 & 56.0 & 56.0 \\
\midrule
\textit{Model based on cross-lingual transfer} \\
CCLM 3M  			& \multirow{2}{*}{\cite{zeng-etal-2023-cross}} & 420 M & 69.6 & 61.6 & 70.5 & 67.8 & 60.3 & 66.0 \\
CCLM 4M  			&  & 970M & 66.8 & 67.2 & 69.9 & 71.7 & 60.4 & 67.2 \\
mBLIP mT0-XL (zero-shot)  	& \multirow{4}{*}{\cite{geigle-etal-2024-mblip}} & 4.9B & 68.1 & 64.8 & 65.9 & 64.9 & 69.7 & 66.7 \\
mBLIP BLOOMZ-7B (zero-shot)  	&  & 8.3B & 57.7 & 56.2 & 59.7 & 59.1 & 60.3 & 58.6 \\
mBLIP mT0-XL (fine-tuned)  	&  & 4.9B & 74.3 & 74.6 & 75.7 & 75.1 & 75.9 & 75.1 \\
mBLIP BLOOMZ-7B (fine-tuned)  	&  & 8.3B & 61.4 & 69.7 & 81.2 & 80.1 & 77.4 & 74.0 \\
CliCoTea  			& \cite{karoui-etal-2023-stop} & 210M & 70.7 & 71.3 & 64.9 & 69.6 & 63.9 & 68.1 \\
Ours Multiling.  			& & \multirow{3}{*}{330M} & 71.9 & 68.1 & 73.7 & 68.6 & 65.9 & 69.5 \\
Ours Biling. 5k  			& & & 69.6 & 64.2 & 68.0 & 67.1 & 60.5 & 65.9 \\
Ours Biling. 25k 			& & & 73.2 & 66.9 & 71.2 & 67.9 & 61.2 & 68.1 \\
\bottomrule
\end{tabular}
}
\end{table}

\end{document}